# SAGC-A68: a space access graph dataset for the classification of spaces and space elements in apartment buildings

Amir Ziaee, Georg Suter.
Design Computing Group, TU Wien, Vienna, Austria
amir.ziaee@tuwien.ac.at

**Abstract.** The analysis of building models for usable area, building safety, and energy use requires accurate classification data of spaces and space elements. To reduce input model preparation effort and errors, automated classification of spaces and space elements is desirable. A barrier hindering the utilization of Graph Deep Learning (GDL) methods to space function and space element classification is a lack of suitable datasets. To bridge this gap, we introduce a dataset, SAGC-A68, which comprises access graphs automatically generated from 68 digital 3D models of space layouts of apartment buildings. This graph-based dataset is well-suited for developing GDL models for space function and space element classification. To demonstrate the potential of the dataset, we employ it to train and evaluate a graph attention network (GAT) that predicts 22 space function and 6 space element classes. The dataset[1] and code[2] used in the experiment are available online.

## 1. Introduction

Building information modeling (BIM) authoring systems are useful for, among other things, detailed space based analysis, such as usable area measurement, building safety, energy, and evacuation path analysis. Classification properties of spaces and related space elements, including doors and windows, are required for these analyses. Usually, classification data needs to be manually entered by users. Manual entry of classification data is problematic because it is prone to errors and could cause inaccurate analysis. Furthermore, manual input data preparation is time-consuming and significantly slows down the analysis process, particularly for large buildings. Thus, automated classification of spaces and space elements is desirable.

To automate classification tasks in BIM and communicate with other authoring software, researchers have investigated machine learning methods, including Graph Deep Learning (GDL, Wang et al., 2022; Buruzs et al., 2022). In GDL, buildings are represented as graphs, with nodes corresponding to objects and edges to logical or spatial relationships between them, respectively. On the one hand, graph data structures are well-suited for representing relationships between objects in a BIM model. Additionally, graphs can be easily queried to extract specific information about elements in the model. On the other hand, graphs can be processed using machine learning techniques. A fundamental challenge is to create suitable datasets to train and evaluate GDLs. In developing such datasets, consideration must be given to the diversity of buildings concerning size, type, functions, or location. In addition, automated graph extraction from source building data is desirable to avoid errors that may exist in manually created graphs. Existing graph-based building datasets are limited because they are not publicly available and lack diversity (Wang et al., 2022; Buruzs et al., 2022). These limitations currently restrict the applicability of GDL methods in the building domain.

[1] https://doi.org/10.5281/zenodo.7805872
[2] https://github.com/A2Amir/SAGC-A68

To address these issues, we present a graph-based building dataset, SAGC-A68, that consists of space access graphs generated automatically from 68 digital 3D models of space layouts of apartment buildings. We use the dataset to train and evaluate an experimental GDL model to classify spaces and space elements. In this study, we follow a workflow which consists of five steps (Eisler and Meyer, 2020). First, we formulate the classification of spaces and space elements in apartment buildings as a node classification problem in space access graphs. Next, we describe a data processing workflow to create our dataset. Space access graphs are pre-processed according to the input requirements for GDLs. We have used the dataset to train and validate a graph attention network (GAT). We present initial results obtained from testing the GAT's capability to predict 22 space function and 6 space element classes.

## 2. Related work

Classification methods for BIM can be categorized into Rule-based, Machine Learning-based (ML), Point Cloud Deep Learning-based (PCDL), Image DL-based (IDL), and Graph DL-based (GDL) methods (Ziaee et al., 2022). GDL refers to a category of machine learning techniques that operate on graph data structures, such as Graph Neural Networks (GNNs, Zhou et al., 2020), Graph Convolutional Networks (GCNs, Kipf et al., 2017) and graph attention networks (GATs, Brody et al., 2021). In GNNs that are applied to buildings, a building is represented as a graph where nodes correspond to objects and edges to logical or spatial relationships between them. Examples for relationships are adjacency or connectivity relationships. Wang, Sacks, and Yeung (2022) developed a space connectivity graph dataset and a GNN to classify nine space functions in apartment buildings by unit. The dataset was created manually from online layout images and consists of 2076 spaces from 224 apartments in three countries. A space connectivity graph models three connection or edge types, namely walls, virtual walls, and doors. Sub-graphs include space adjacency and space access graphs.

Buruzs et al. (2022) created a dataset for classifying space functions in apartment buildings by floor. The dataset comprises graphs that represent space connections by door, opening, stair, or touching spaces. A geometric algorithm was used to detect IfcSpace entities and to inject them into IFC models. Another geometric algorithm was used to create a space access graph dataset. The dataset consists of 1500 spaces with eight space functions and was derived from 15 online IFC models. The dataset was then utilized for training and testing a GCN to classify space functions.

A limitation of existing studies is that their datasets lack diversity in terms of size, type, functions, or location. In addition, graph creation needs to be automated such that larger, high data quality datasets can be developed.

Table 1: Comparing different graph datasets.

| Dataset | Space layout scope | Classification task | Number of classes | Number of instances | Graph creation | Countries | Source data |
|---|---|---|---|---|---|---|---|
| Wang et al, 2022 | Unit | Space function | 9 | 2076 | Manual | 3 | Online images |
| Buruzs et al, 2022 | Floor | Space function | 8 | 1500 | Automated | - | IFC models |
| **Our dataset** | **Floor** | **Space function** and **Space element** | **22** | **2426** | **Automated** | **13** | BIM/CAD models |
| | | | **6** | **2445** | | | |

Existing work indicates a significant potential for graph methods to address classification needs in BIM in general and space classification in particular. However, there are still few studies, and the potential of GDL methods has not been explored fully. Our SAGC-A68 dataset aims to address a lack of publicly available graph-based building datasets (Table 1, Appendix A.1). Space access graphs in SAGC-A68 were automatically generated from BIM/CAD models of apartment buildings in 13 countries. We use the dataset to train and evaluate an extended GAT that predicts space function and space element classes in space access graphs.

## 3. Problem formulation

We aim to classify space and space element nodes in a space access graph of an apartment building floor. Each node represents an internal space, a loggia, an access balcony, a door, or an opening. Each space access edge connects a door or opening node with a space node. Given an unlabeled space access graph, the task is to assign a space function or space element class label to each node. Examples of space function classes are 'LivingRoom', 'Kitchen', or 'Elevator', and of space element classes 'InternalDoor' or 'UnitDoor'. Distinguishing door classes is relevant for space-based applications, such as evacuation path analysis (e.g., OIB, 2015). We aim to develop a GAT for classifying all space access graph nodes (Figure 1). GATs are GNN models that have been shown to achieve higher performance in diverse benchmark machine learning tasks than other GNN architectures, such as GCN and GraphSAGE (Brody et al., 2021).

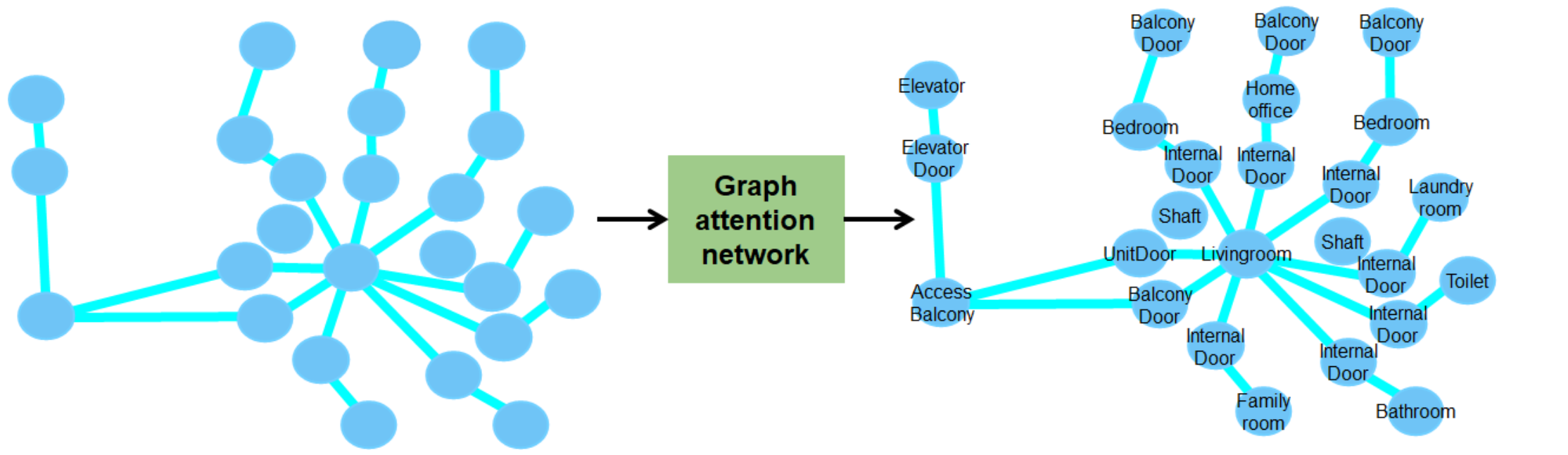

(a) Input graph (b) Graph attention network (c) Labeled output graph

Figure 1: Classification of space and space elements.

Space function and space element classes in apartment buildings that our GAT model classifies are shown in bold type in Table 2. Circulation spaces, such as stairways or elevators, are classified because our aim is to classify nodes of space access graphs of entire floors instead of individual apartment units. We have identified 22 space and 6 space element classes for the apartment buildings in the SAGC-A68 dataset.

Table 2: Space and space element class hierarchies and instance counts in the SAGC-A668 dataset. Predicted classes are shown in bold type.

| Space function classes | | Space element classes | |
|---|---|---|---|
| **Name** | **Count** | **Name** | **Count** |
| Space | | SpaceElement | |
| ResidentialSpace | | SpaceEnclosingElement | |
| CommunalSpace | | **Opening** | **140** |
| **DiningRoom** | **3** | Door | |
| **FamilyRoom** | **6** | **InternalDoor** | **1428** |
| **LivingRoom** | **275** | **UnitDoor** | **291** |
| PrivateSpace | | **SideEntranceDoor** | **10** |
| **Bedroom** | **495** | **ElevatorDoor** | **84** |
| **MasterBedroom** | **23** | **BalconyDoor** | **492** |
| **BoxRoom** | **2** | | |
| **HomeOffice** | **8** | | |
| ServiceSpace | | | |
| **Shaft** | **403** | | |
| **StorageRoom** | **84** | | |
| **WalkInCloset** | **2** | | |
| SanitarySpace | | | |
| **Bathroom** | **274** | | |
| **Toilet** | **145** | | |
| **Kitchen** | **117** | | |
| **LaundryRoom** | **57** | | |
| CirculationSpace | | | |
| VerticalCirculationSpace | | | |
| **Elevator** | **86** | | |
| **Stairway** | **70** | | |
| HorizontalCirculationSpace | | | |
| **Entrance** | **67** | | |
| **Hallway** | **12** | | |
| **MainHallway** | **18** | | |
| **InternalHallway** | **152** | | |
| External | | | |
| **AccessBalcony** | **19** | | |
| **Loggia** | **108** | | |

## 4. Data collection

The SAGC-A68 dataset is created from the same source building data that we used to create the SFS-A68 dataset for space function segmentation (Ziaee et al., 2022). It comprises 68 digital 3D models of space layouts of apartment buildings designed or built in the years 1952-2019. In total, source data includes 275 apartments. 78% of the models are from buildings in Austria, Germany, and Switzerland, and 22% from 10 countries in Europe, the USA, and Southern America, respectively.

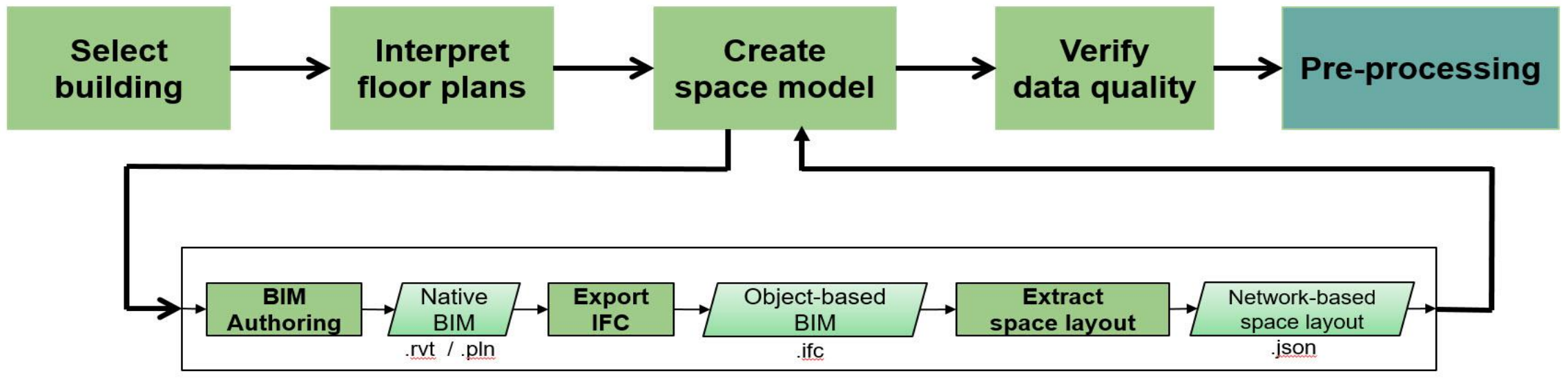

Figure 2: Data collection workflow, adapted from Ziaee and Suter (2022) and Suter (2022).

We have adapted the existing data collection workflow for the SFS-A68 dataset to create the SAGC-A68 dataset (Figure 2). A detailed description of the data collection workflow is given in Ziaee and Suter (2022) and Suter (2022). Space access graphs are extracted from space layouts and saved as lists of spaces, space elements, and space access edges in the JavaScript Object Notation (JSON) file format (JSON, 1999). Space features include space function class, center point, the axis-aligned volume bounding box, gross floor area, volume, door/opening count, and window count. Space element features include space element class, center point, width, height, axis-aligned face bounding box, and face area. Space access edge features include length, elevation difference between related nodes, and edge angle relative to the xy-plane.

## 5. Pre-processing

Since the generated space access graphs were written in JavaScript Object Notation (JSON) format, their node and edge list representations plus space, space element, and space access edge features were converted to matrix representations and passed to the NetworkX network analysis library (Hagberg et al., 2008) to compute additional graph-based features that provide centrality and clustering measures for space access graph nodes and edges. We computed five graph-based node features: node degree, betweenness centrality (Brandes et al., 2001), page rank (Langville et al., 2005), closeness centrality (Freeman et al., 2002), degree centrality, and clustering coefficient (Saramäki et al., 2007). In previous work, graph-based features are limited to node degree and closeness centrality (Wang, Sacks, and Yeung, 2022). Space access edges have two additional features: edge betweenness centrality (Brandes et al., 2001) and the angle between an edge and the local coordinate system x-axis.

Lastly, the position of each node was normalized based on the bounding box of each space access graph, and a scaling method known as standardization was used to shift each node and edge feature distribution to have a mean of zero and a standard deviation of one unit (Ziaee et al., 2022).

## 6. Model construction

Applying the pre-processing workflow resulted in 68 space access graphs, each including nodes with feature vectors of length 20 and edge vectors of length 5. Since the dataset used to train a model must be different from the one used to evaluate its performance, the space access graphs were split into a training and a testing dataset with a ratio of 90% (61 graphs) to 10% (7 graphs). We did not perform hyperparameter fine-tuning in this study, which requires a validation dataset. Thus, the training dataset is used to train the GAT and the test dataset is used to evaluate it.

A GAT can learn useful representations of nodes by utilizing the graph structure (Brody et al., 2021). It achieves this by an attention mechanism that weights the contribution of each neighboring node to the final embedding, making it capable of identifying salient features in the graph. It assigns different attention weights to neighboring nodes and then computes a weighted average of the neighbor embeddings using these attention weights. This way, more important neighbors contribute more to the final node embedding, and less important neighbors have less influence. Since a typical GAT only takes node features into node embeddings, we extend GAT to incorporate edge features. Since space access graphs are geometric networks, space access edge lengths correspond to the Euclidean distance between related nodes. Thus in our extended GAT we can use edge length as a feature. In future work, this extension allows us to consider additional space connectivity relationships, such as the space adjacency relationship (Wang et al., 2022). More details on our extended GAT are provided in Appendix A.2.

To train the extended GAT from scratch, we used a simple training strategy, starting with a learning rate of 0.001, training the model 5000 epochs, and saving the best model with the lowest error on the training dataset. During the network's training, an Adam optimizer (Zhang, 2018) was used to minimize a developed version of focal loss generalized to the multi-class case (Lin et al., 2017). This loss is essentially an enhancement to cross-entropy loss and is useful for classification tasks when there is a large class imbalance in terms of the number of instances per class. The class imbalance is due to the inherent nature of the architectural design and construction process. Spaces, such as 'Bedroom' or 'LivingRoom', tend to have more numbers compared to other spaces, such as 'WalkInCloset' or 'LaundryRoom'.

## 7. Model validation

The outputs of the extended GAT model for the test dataset are evaluated by Precision, Recall, and F1-Score metrics. Precision quantifies the proportion of correct positive predictions made based on Equation 1:

$$Precision = \frac{TP}{TP + FP} \qquad (1)$$

where TP is the number of true-positives, and FP is the number of false-positives. Recall computes the ratio between the number of positives correctly classified as positive to the total number of positive samples based on Equation 2:

$$Recall = \frac{TP}{TP + FN} \qquad (2)$$

where FN is the number of false-negatives. F1-score sums up the predictive performance of a model by combining precision and recall metrics based on Equation 3:

$$F1 - Score = 2 * \left(\frac{Precision * Recall}{Precision + Recall}\right) \qquad (3)$$

Evaluation results for the test dataset are given in Table 3. Metrics are computed only for classes that are present in the test dataset. For example, 'HomeOffice' and 'SideEntranceDoor' are not present in any test space access graph, and therefore no results are reported for these classes.

Table 3: Precision, Recall and F1-Score for the test dataset.

| **Space function class** | **Training instance count** | **Test instance count** | **Precision** | **Recall** | **F1-Score** |
|---|---|---|---|---|---|
| AccessBalcony | 17 | 2 | 0.00 | 0.00 | 0.00 |
| Bathroom | 241 | 33 | 0.64 | 0.76 | 0.69 |
| Bedroom | 434 | 61 | 0.84 | 0.59 | 0.69 |
| BoxRoom | 2 | 0 | - | - | - |
| DiningRoom | 2 | 1 | 0.00 | 0.00 | 0.00 |
| Elevator | 80 | 6 | 0.30 | 0.50 | 0.37 |
| Entrance | 63 | 4 | 0.00 | 0.00 | 0.00 |
| FamilyRoom | 6 | 0 | - | - | - |
| HomeOffice | 8 | 0 | - | - | - |
| Hallway | 12 | 0 | - | - | - |
| InternalHallway | 136 | 16 | 0.75 | 0.94 | 0.83 |
| Kitchen | 112 | 5 | 0.15 | 0.60 | 0.24 |
| LaundryRoom | 47 | 10 | 0.33 | 0.10 | 0.15 |
| LivingRoom | 244 | 31 | 0.93 | 0.81 | 0.86 |
| Loggia | 103 | 5 | 1.00 | 0.80 | 0.89 |
| MasterBedroom | 22 | 1 | 0.00 | 0.00 | 0.00 |
| MainHallway | 18 | 0 | - | - | - |
| Shaft | 358 | 45 | 0.92 | 0.98 | 0.95 |
| Stairway | 65 | 5 | 0.67 | 0.80 | 0.73 |
| StorageRoom | 71 | 13 | 0.50 | 0.31 | 0.38 |
| Toilet | 129 | 16 | 0.55 | 0.75 | 0.63 |
| WalkInCloset | 2 | 0 | - | - | - |
| **Space element class** | | | | | |
| BalconyDoor | 404 | 88 | 1.00 | 0.68 | 0.81 |
| ElevatorDoor | 78 | 6 | 0.29 | 0.33 | 0.31 |
| InternalDoor | 1267 | 161 | 0.82 | 0.94 | 0.88 |
| Opening | 134 | 6 | 0.86 | 1.00 | 0.92 |
| SideEntranceDoor | 10 | 0 | - | - | - |
| UnitDoor | 259 | 32 | 0.81 | 0.81 | 0.81 |
| **Total count/Weighted Avg** | **4324** | **547** | **0.80** | **0.77** | **0.77** |

Figure 3 presents the confusion matrix for the recall metric in a normalized percentage format.

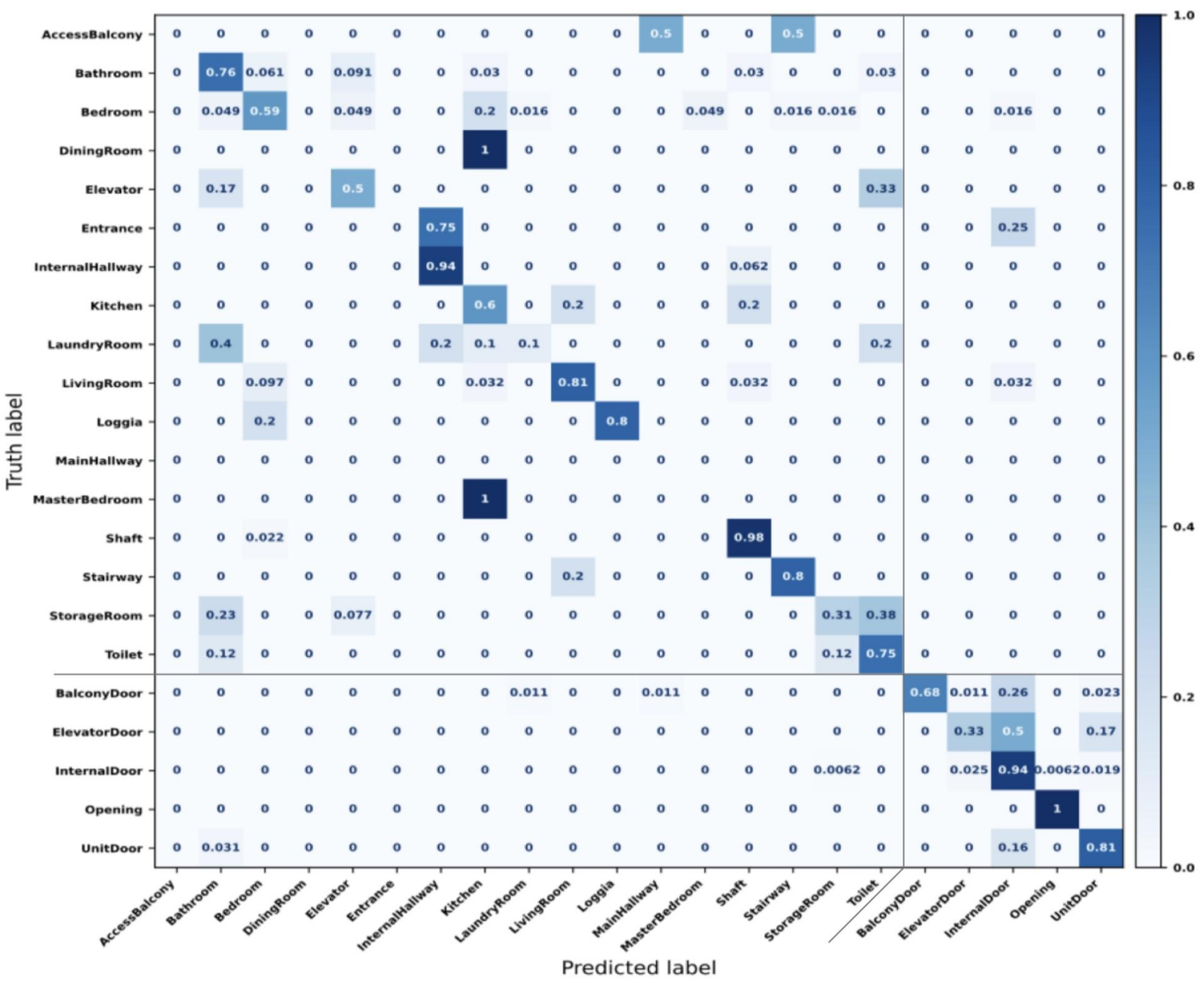

Figure 3: Confusion matrix for the test dataset.

Metrics are computed only for classes that are present in the test dataset.

Embeddings of the test dataset from the penultimate layer of the model are extracted and the T-distributed Stochastic Neighborhood Embedding (tSNE) method (Linderman et al., 2017) is used for reducing the dimensionality into two-dimensional space (Figure 4).

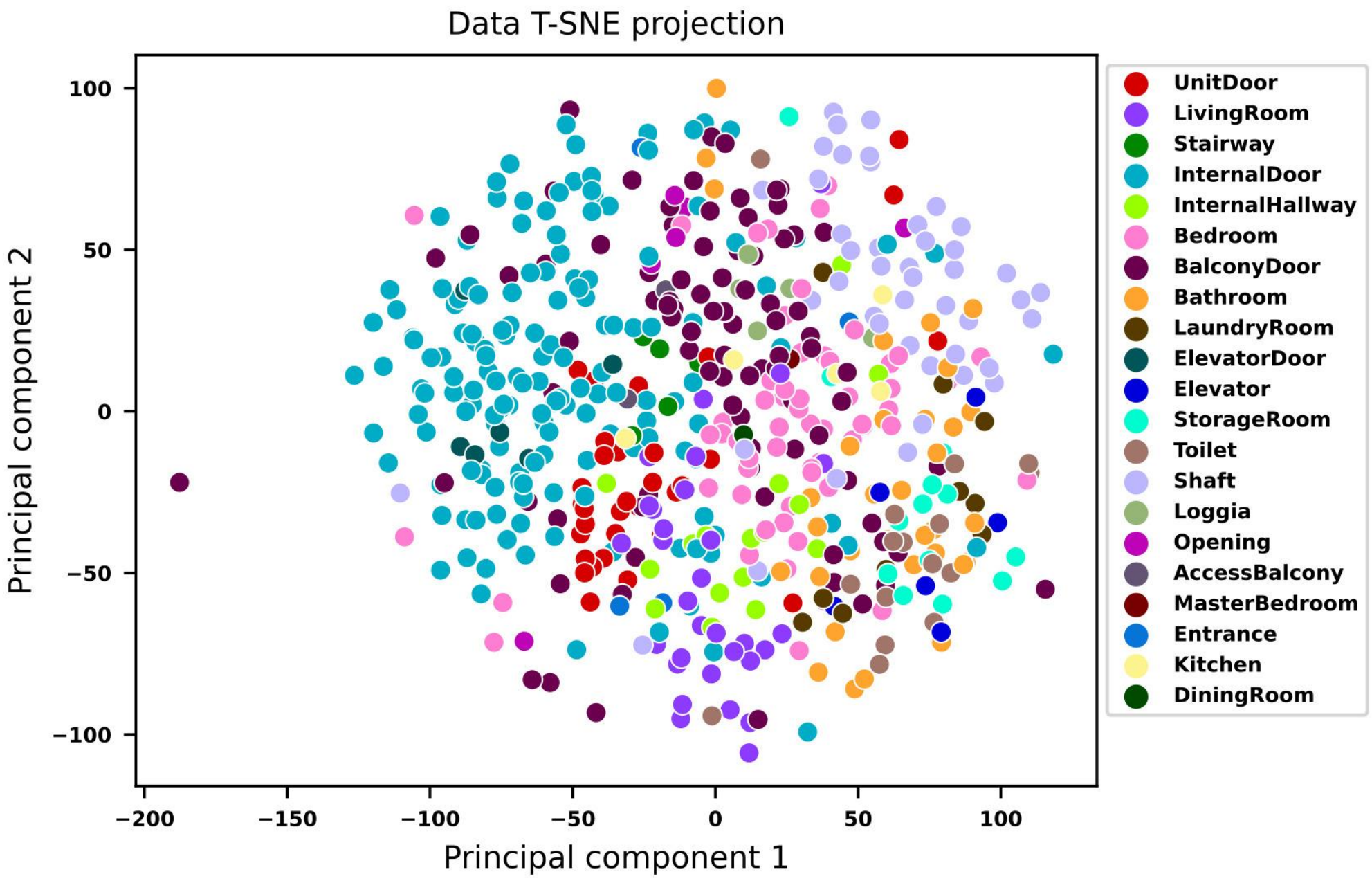

Figure 4: Dimensionality reduction of the test embeddings from the penultimate layer of the model.

## 8. Discussion

Our GAT model made correct predictions for all classes except ‘AccessBalcony’, ‘DiningRoom’, and ‘Entrance’ in the test dataset (Table 3, Figure 3). For the latter classes, no predictions were made, probably due to a lack of training data. The model could generally well distinguish spaces from space elements (Figures 3 and 4). Except for 'ElevatorDoor', the model achieved a high prediction performance (F1-Score between 0.81 and 0.92) for doors and 'Opening'.

Prediction performance varies significantly across space function classes (Table 3). This finding is consistent with results from previous studies (Wang, Sacks, and Yeung, 2022; Buruzs et al., 2022). Three performance groups can be identified. Performance in the first group, which includes 'Shaft', 'Loggia', 'LivingRoom', and 'InternalHallway', is found to be high (that is, all metrics are equal or greater than 0.75). Compared with other spaces, these spaces usually have a higher ('LivingRoom', 'InternalHallway') or lower ('Shaft', 'Loggia') number of incident edges in space access graphs. The next group, which includes 'Stairway', 'Bathroom', 'Bedroom', and 'Toilet', is found to have intermediate performance (F1-Score between 0.63 and 0.73). 'Bedroom' has higher precision but lower recall. For example, 20% of actual 'Bedrooms' were misclassified as 'Kitchens' (Figure 3). Finally, performance was low (F1-Score between 0.15 and 0.38) for 'StorageRoom', 'Elevator', 'Kitchen', and 'LaundryRoom'. These spaces are similar in size and may need more distinct features to improve their performance. It should be noted that only closed kitchens are modeled in the SAGC-A68 dataset. Open kitchens are considered part of living rooms. We plan to address this issue in future releases of the dataset.

The uneven prediction performance of our GAT model may be attributed in part to a class imbalance between spaces and space elements in the SAGC-A68 dataset. There are 22 space classes among 2426 spaces, resulting in an average of approximately 110 instances per space class. On the other hand, there are 6 space element classes among 2445 space elements, resulting in an average of approximately 408 instances per space element class. Therefore, we can observe a class imbalance where the space element classes have, on average, a significantly higher number of instances per class compared to the space classes.

In our previous study, the space function segmentation network model resulted in similarly uneven prediction performance (Ziaee and Suter, 2022). However, certain classes have different performances in segmentation and GAT models. For example, 'Shafts' have low performance in the segmentation model due to small floor areas. Since they are usually inaccessible, they are more easily detected by the GAT model.

## 9. Conclusion

We presented a space access graph dataset for classifying space functions and space elements in apartment buildings. We adapted an existing data processing workflow to automatically extract space access graphs from CAD or BIM source space data. We plan to extend the dataset to improve its diversity, particularly concerning location. Currently, most buildings in the dataset are located in Europe. We plan to investigate how our GAT model's prediction performance could be improved, e.g., by integrating space adjacency graphs or augmenting node or edge features. We further aim to evaluate and compare the efficacy of alternative GNN model architectures. Varying prediction performances for certain classes in our previously developed space function network segmentation model and the GAT model presented in this paper point towards combining these models into an ensemble learning model with improved performance.

## Acknowledgments

The authors gratefully acknowledge support by Grant Austrian Science Fund (FWF): I 5171-N, Laura Keiblinger, and participants in course '259.428-2021S Architectural Morphology' at TU Wien for data collection.

## References

Buruzs, A., Šipetić, M., Blank-Landeshammer, B., & Zucker, G. (2022). IFC BIM Model Enrichment with Space Function Information Using Graph Neural Networks. Energies, 15(8), 2937 %@ 1996-1073.https://doi.org/10.3390/en15082937

Brody, S., Alon, U., & Yahav, E. (2021). How attentive are graph attention networks?. arXiv preprint arXiv:2105.14491. https://doi.org/10.48550/arXiv.2105.14491

Brandes, U. (2001). A faster algorithm for betweenness centrality. Journal of mathematical sociology, 25(2), 163-177. https://doi.org/10.1080/0022250X.2001.9990249

Eisler, S. and Meyer, J. (2020). Visual Analytics and Human Involvement in Machine Learning. arXiv preprint arXiv:2005.06057. https://doi.org/10.48550/arXiv.2005.06057

Freeman, L. C. (2002). Centrality in social networks: Conceptual clarification. Social network: critical concepts in sociology. Londres: Routledge, 1, 238-263. https://doi.org/10.1016/0378-8733%2878%2990021-7

Hagberg, A., Swart, P., & S Chult, D. (2008). Exploring network structure, dynamics, and function using NetworkX (No. LA-UR-08-05495; LA-UR-08-5495). Los Alamos National Lab.(LANL), Los Alamos, NM (United States).

JSON, JavaScript Object Notation (1999). The JSON data interchange syntax. [Online] Available: https://www.json.org/json-en.html [Accessed 7 April 2023]

Kipf, T.N.; Welling, M. (2017). Semi-Supervised Classification with Graph Convolutional Networks. arXiv 2017, arXiv:1609.02907. https://doi.org/10.1007/s11063-021-10487-w

Lin, T. Y., Goyal, P., Girshick, R., He, K., & Dollár, P. (2017). Focal loss for dense object detection. In Proceedings of the IEEE international conference on computer vision (pp. 2980-2988). http://dx.doi.org/10.1109/TPAMI.2018.2858826

Linderman, G. C., Rachh, M., Hoskins, J. G., Steinerberger, S., & Kluger, Y. (2017). Efficient algorithms for t-distributed stochastic neighborhood embedding. arXiv preprint arXiv:1712.09005. https://doi.org/10.1038/s41592-018-0308-4

Langville, A. N., & Meyer, C. D. (2005). A survey of eigenvector methods for web information retrieval. SIAM review, 47(1), 135-161. https://doi.org/10.1137/S0036144503424786

OIB-330.2-011/15: OIB-Richtlinie 2. (2015). Brandschutz, Richtlinien des Österreichischen Instituts für Bautechnik. [Online] Available: https://www.oib.or.at/sites/default/files/richtlinie_2_26.03.15.pdf [Accessed 6 April 2023]

Suter, G. (2022). Modeling multiple space views for schematic building design using space ontologies and layout transformation operations. Automation in Construction, 134, 104041. https://doi.org/10.1016/j.autcon.2021.104041

Saramäki, J., Kivelä, M., Onnela, J. P., Kaski, K., & Kertesz, J. (2007). Generalizations of the clustering coefficient to weighted complex networks. Physical Review E, 75(2), 027105. https://doi.org/10.48550/arXiv.cond-mat/0608670

Wang, Z., Sacks, R. and Yeung, T. (2022). Exploring graph neural networks for semantic enrichment: Room type classification. Automation in Construction, 134, 104039. https://doi.org/10.1016/j.autcon.2021.104039

Wu, L., Cui, P., Pei, J., Zhao, L., & Song, L. (2022). Graph neural networks (pp. 27-37). Springer Singapore. https://doi.org/10.1007/978-981-16-6054-2_3

Ziaee, A., Suter, G. and Barada, M. (2022). SFS-A68: a dataset for the segmentation of space functions in apartment buildings. [Online] Available: https://doi.org/10.5281/zenodo.6426871 [Accessed 6 April 2023]

Ziaee, Amir & Suter, Georg. (2022). SFS-A68: a Dataset for the Segmentation of Space Functions in Apartment Buildings. http://dx.doi.org/10.7146/aul.455.c222

Zhou, J., Cui, G., Hu, S., Zhang, Z., Yang, C., Liu, Z., ... & Sun, M. (2020). Graph neural networks: A review of methods and applications. AI open, 1, 57-81. https://doi.org/10.1016/j.aiopen.2021.01.001

Ziaee, A., & Çano, E. (2022). Batch Layer Normalization, A new normalization layer for CNNs and RNN. arXiv preprint arXiv:2209.08898. https://doi.org/10.1145/3571560.3571566

Zhang, Z. (2018) Improved Adam optimizer for deep neural networks. IEEE/ACM 26th International Symposium on Quality of Service (IWQoS), 2018. IEEE, 1-2. https://doi.org/10.1109/IWQoS.2018.8624183

Ziaee, Amir, Georg, Suter, & Keiblinger, Laura. (2023). SAGC-A68: A space access graph dataset for the classification of spaces and space elements in apartment buildings (1.0.1) [Data set]. 30th International Workshop on Intelligent Computing in Engineering (EG-ICE), London, UK. Zenodo. https://doi.org/10.5281/zenodo.7805872

### Appendix A.1 Space access graph features

Each node in a space access graph in SAGC-A68 dataset that was used to construct the GAT model is characterized by 15 features and 5 derived graph-based features (Section 5). These include Center point, Width, Height, Depth, Area, Volume, Is_internal, Door_opening_quantity, Window_quantity, Max_door_width, and Class label. Similarly, each edge is characterized by three features and 2 derived graph-based features (Section 5). These include Z_angle, Delta_z, and Length. The dataset comprises a total of 4871 nodes and 4566 edges, with the label nodes classified into 28 categories (Table 2).

Figure A.1 shows examples of space access graphs with the corresponding node labels.

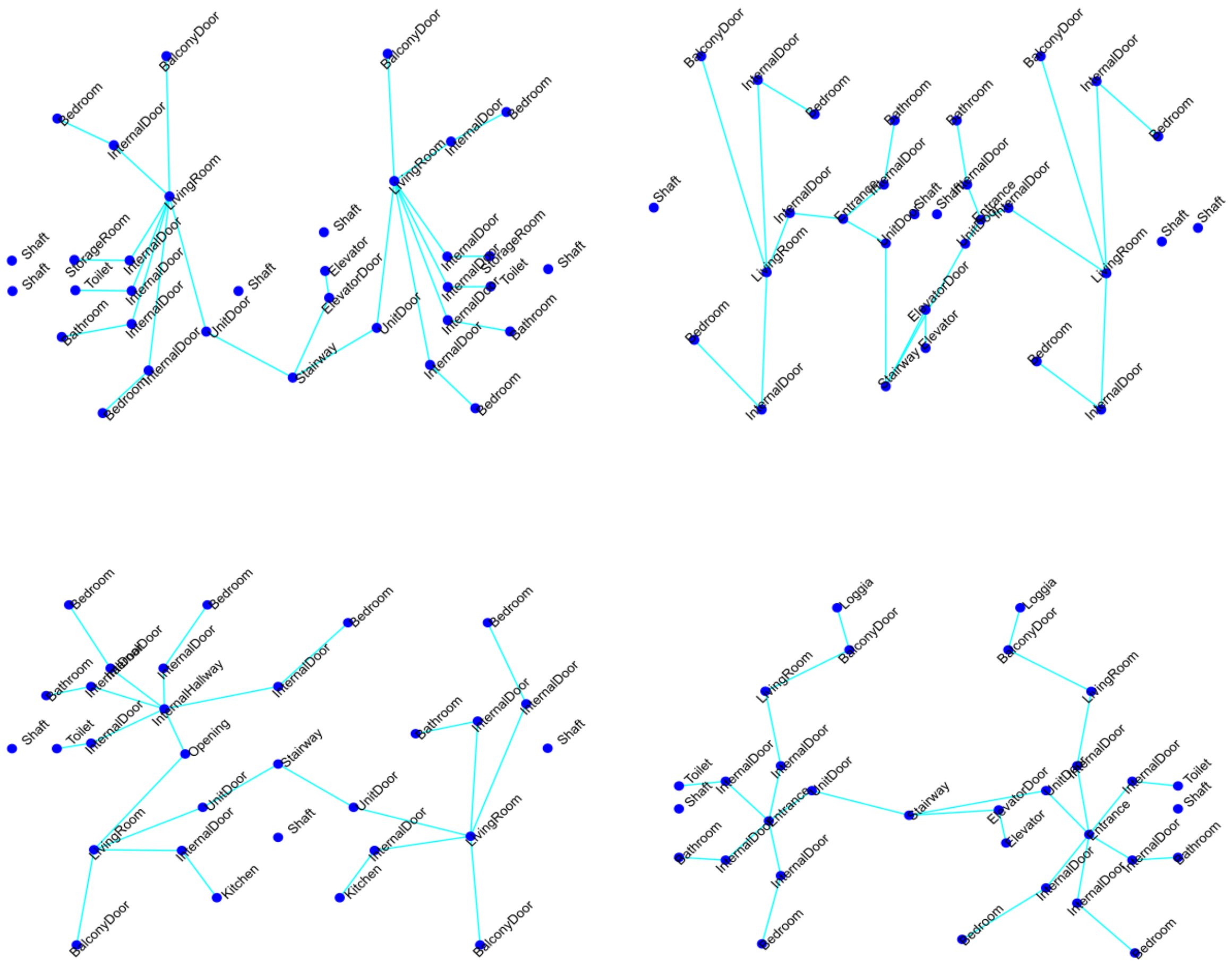

Figure A1. Example space access graphs.

### A.2. Extended graph attention network

We extended the GAT (Brody et al., 2021) by first computing node attentions (Step 1), compute corresponding edge embeddings (Step 2) and aggregating and applying a softmax function to generate a probability distribution 'alpha' for the neighbors of a final node. The attention coefficient 'alpha' is then utilized to compute a weighted sum of neighbors and obtain a new representation for the final node (Step 3).

In our extended GAT model, the number of attention heads ($N_{heads}$) is set sequentially to 3, 2, 1, and $N_{out}$, where $N_{out}$ is equal to 28, the number of space and space element classes.

---

Extended graph attention network layer.

---

**Input:** Let $G = (N, E)$ be the input graph with node features h and edge features k, where N is the set of nodes and E is the set of edges. Let $h$ be the node feature matrix of shape $(N, N_{in})$, where $N_{in}$ is the number of input node features, $k$ be the edge feature matrix of shape $(E, E_{in})$, where $E_{in}$ is the number of input edge features, and $N_{heads}$ be the number of attention heads.

**Output:** Let $h_i^{(l+1)}$ be the node output matrix of shape $(N, N_{heads} * N_{heads} * N_{out})$, where $N_{out}$ is the number of output node features.

**Step 1.** Node attention mechanism (Brody et al., 2021) can be presented as:

$$h_i'^{(l)} = \sum_{j \in \mathcal{N}(i)} \alpha_{ij}^{(l)} W^{(l)} h_j^{(l)}$$

$$\alpha_{ij}^{(l)} = \mathrm{softmax}_{\mathrm{i}}(e_{ij}^{(l)})$$

$$e_{ij}^{(l)} = \vec{a}^{T(l)} \mathrm{LeakyReLU}\left(W^{(l)} h_i + W^{(l)} h_j\right),$$

where $h_j^{(l)}$ is the representation of source node $j$ at layer $l$, $W^{(l)}$ is the weight matrix of shape $(N_{in}, N_{heads} * N_{heads} * N_{out})$, $\alpha_{i,j}^{(l)}$ is the attention coefficient between nodes $i$ and $j$ at layer $l$, $N(i)$ is the set of neighbor nodes of $i$, $\vec{a}^{(l)}$ is a learnable attention parameter vector, and $h_i'^{(l)}$ is new the representation of target node $i$ of the shape $(N, N_{heads}, N_{heads} * N_{out})$ at layer $l$.

**Step 2.** Edge feature embedding can be presented as:

$$k_i'^{(l)} = W^{(l)} k_i^{(l)},$$

where $k_i^{(l)}$ is the representation of edge $i$ at layer $l$, $W^{(l)}$ is the weight matrix of shape $(E_{in}, N_{heads} * N_{out})$, and $k_i'^{(l)}$ is new the representation of edge $i$ of the shape $(E, N_{heads} * N_{out})$ broadcasted to the shape of $(E, N_{heads}, N_{heads} * N_{out})$ at layer $l$.

**Step 3.** Update and Aggregation mechanism can be presented as:

$$h_i^{(l+1)} = \sum_{j \in \mathcal{N}(i)} \alpha_{ij}^{(l)} m_{ij}^{(l)}$$

$$\alpha_{ij}^{(l)} = \mathrm{softmax}_{\mathrm{j}}(m_{ij}^{(l)})$$

$$m_{ij}^{(l)} = h_j'^{(l)} + k_{ij}'^{(l)},$$

where $\mathrm{m}_{ij}^{(l)}$ is an intermediate message that encodes information from both the source node and the edge, $h_i^{(l+1)}$ is new the representation of target node $i$ of the shape $(N, N_{heads}, N_{heads} * N_{out})$ at layer $l+1$ reshaped to the shape $(N, N_{heads} * N_{heads} * N_{out})$.

---